%% file: main.tex
\documentclass[
  aps,
  prl,
  twocolumn,
  10pt,
  balancelastpage,
  amsmath,
  amssymb,
  nofootinbib,
  nobibnotes,
  floatfix,
  superscriptaddress
]{revtex4-2}

\input{preamble}

\begin{document}
\title{Cognitive Maps in Large Language Models}

\author{Albert Qin}
\email{aq8221@princeton.edu}
\affiliation{Department of Physics, Princeton University, Princeton, New Jersey}

\author{Jonathan D. Cohen}
\affiliation{Princeton Neuroscience Institute, Princeton University, Princeton, New Jersey}

\author{Gautam Reddy}
\email{greddy@princeton.edu}
\affiliation{Department of Physics, Princeton University, Princeton, New Jersey}

\begin{abstract}
    \input{manuscript/sections/abstract}
\end{abstract}

\maketitle

\section{Introduction}
\input{manuscript/sections/introduction}

\section{Results}
\input{manuscript/figures/fig_1_caption}
\input{manuscript/sections/results_01_behavior}
\input{manuscript/sections/results_02_transformers}
\input{manuscript/figures/fig_2_caption}
\input{manuscript/sections/results_03_separability}
\input{manuscript/figures/fig_3_caption}
\input{manuscript/sections/results_04_linearity}
\input{manuscript/figures/fig_4_caption}
\input{manuscript/sections/results_05_prospective_retrospective}
\input{manuscript/figures/fig_5_caption}
\input{manuscript/sections/results_06_key_value_binding}

\section{Discussion}
\input{manuscript/sections/discussion}

\section*{Code Availability}
Code and data for reproducing the results presented in this article
are available at
\weblink{https://github.com/albertqin8221/cognitive-maps-in-llms/releases}.

\section*{Acknowledgments}
We thank Tim Buschman, Declan Campbell, Tim Lin, Keith Murray, as well as members of the Reddy Lab for helpful discussions. AQ was supported by a fellowship from the Fannie and John Hertz Foundation. JDC was supported by a Vannevar Bush Faculty Fellowship from the Office of Naval Research. GR was partially supported by a joint research agreement between NTT Research Inc. and Princeton University, a grant from Coefficient Giving, and a seed grant from the Princeton AI Lab. The analyses presented in this article were performed on Della, a computing cluster managed and supported by Princeton Research Computing, a consortium of groups including the Princeton Institute for Computational Science and Engineering (PICSciE) and Research Computing at Princeton University.

\section*{Author contributions}
A.Q. and G.R. conceived the study. J.D.C. contributed to its conceptual development. A.Q. developed the software, performed the experiments, and analyzed the results. All authors contributed to writing the manuscript.

\section*{Competing interests}
The authors declare that they have no competing interests.

\input{manuscript/references}
\input{supplementary/supplement}
\end{document}

%% file: manuscript/sections/abstract.tex
It has long been proposed that animals construct cognitive maps to support flexible behavior in novel environments. We tested whether large language models form analogous internal representations using relational reasoning tasks and causal ``mosaic'' experiments. Models beyond the ten-billion-parameter scale inferred unstated relationships and generalized to synthetic entity names. Across models and tasks, models formed abstract representations of relational structure, additively combining them with representations of entity identity. Relationships were primarily stored as pairwise links and composed retrospectively, rather than prospectively encoded in a global coordinate system. A sparse, distributed set of key--value pairs mediated binding; perturbing them implanted ``false memories'' that produced predictable behavioral changes. Our findings reveal a conserved organization analogous to a cognitive map supporting flexible and systematic reasoning in language models.

%% file: manuscript/sections/introduction.tex
A hallmark of cognition is the ability to identify structure and make inferences from limited experience. A classic example of this ability is observed during spatial navigation: with sufficient experience in a maze, animals can plan a novel route to reach a goal \cite{tolman_1948_cognitive_maps}. In humans, this ability is observed across virtually all domains that involve relational reasoning; for example, if we are told that Alice’s daughter is Beth and Beth’s daughter is Carol, we can infer that Alice is Carol’s grandmother. Such abilities are made possible by the fact that certain relational structures recur across many contexts: navigation on a flat terrain must obey the rules of two-dimensional geometry, and many human families can be organized into similar kinship structures. It has long been proposed that animals construct internal representations---variously referred to as schemata \cite{bartlett_1932_remembering}, mental models \cite{craik_1943_nature_of_explanation}, or \textit{cognitive maps} \cite{tolman_1948_cognitive_maps, okeefe_1978_hippocampal_map}---that capture such underlying structure to support flexible behavior beyond direct experience.
 
To construct such a representation, a learner must combine previously acquired knowledge about the relevant relational structure (e.g., the abstract structure of a family tree) with information specific to the current context (e.g., the names ``Alice'', ``Beth'', and ``Carol''). Several lines of theoretical work, spanning memory \cite{whittington_2020_tolman_eichenbaum_machine, chandra_2025_vectorhash, kymn2024binding}, structure learning \cite{kemp_2008_structural_form, rumelhart_1993_connectionist_representations, rumelhart_1980_schemata, george2021clone}, and abstract reasoning \cite{webb_2021_emergent_symbols, webb_2024_relational_bottleneck, griffiths_2026_neural_symbols}, can be understood as addressing complementary aspects of this problem. Although their details differ, many of these theoretical models share the feature that context-specific information is bound to a separable and reusable (i.e., abstract) representation of relational structure. This organization allows previously acquired structural knowledge to be rapidly applied to the particulars of a new context. Therefore, to the extent that these theories account for aspects of human cognition and patterns in neural and behavioral data, they suggest that animal brains construct cognitive maps whose organization reflects the recurrence of relational structure across contexts.

This hypothesis immediately raises the question of whether analogous representational strategies can arise in a different system that learns from data with similar statistical features. Modern large language models provide a clear example of such a system. These models exhibit strikingly broad capabilities that arise through training on large natural language corpora \cite{brown_2020_few_shot, wei_2022_emergent, srivastava_2023_bigbench}. Crucially, these corpora contain recurring semantic structures that are instantiated by different entities in different contexts \cite{baker_1998_framenet, chambers_2009_narrative_schemas}, which may in turn arise from the recurrence of certain relational structures across the situations that natural language describes. Therefore, learning from these corpora using objectives that favor generalization across contexts (e.g., next-token prediction) may promote representational strategies that capture recurring relational structure. Crucially, such strategies are not specified a priori but rather emerge through training; their particular form must therefore be characterized empirically. 

Here, we studied the behavior and internal representations of trained large language models when asked to perform a set of relational-reasoning tasks. We found that models could infer unstated relationships and retained this ability when familiar entity names were replaced with novel ones (Fig. \figref{fig:behavior}). By perturbing the models’ internal states, we found evidence for an abstract representation of relational structure separable from the representation of entity identity (Fig. \figref{fig:separability}). These two representations combined additively (Fig. \figref{fig:additivity}). Multi-step inference relied primarily on separately represented relationships that were composed when required by a query (Fig. \figref{fig:composition}). Finally, two separate, sparse sets of key--value pairs supported retrieval of structural information from entity names and vice versa (Fig. \figref{fig:binding}). The general structure of the models' learned representations resembles that of a classical symbol-processing system (Fig. \figref{fig:cognitive-map}). Taken together, our findings reveal how language models organize information in their internal states to support flexible and systematic relational reasoning.

%% file: manuscript/figures/fig_1_caption.tex
\begin{figure*}[t]
    \centering
    \includegraphics[width=\textwidth]{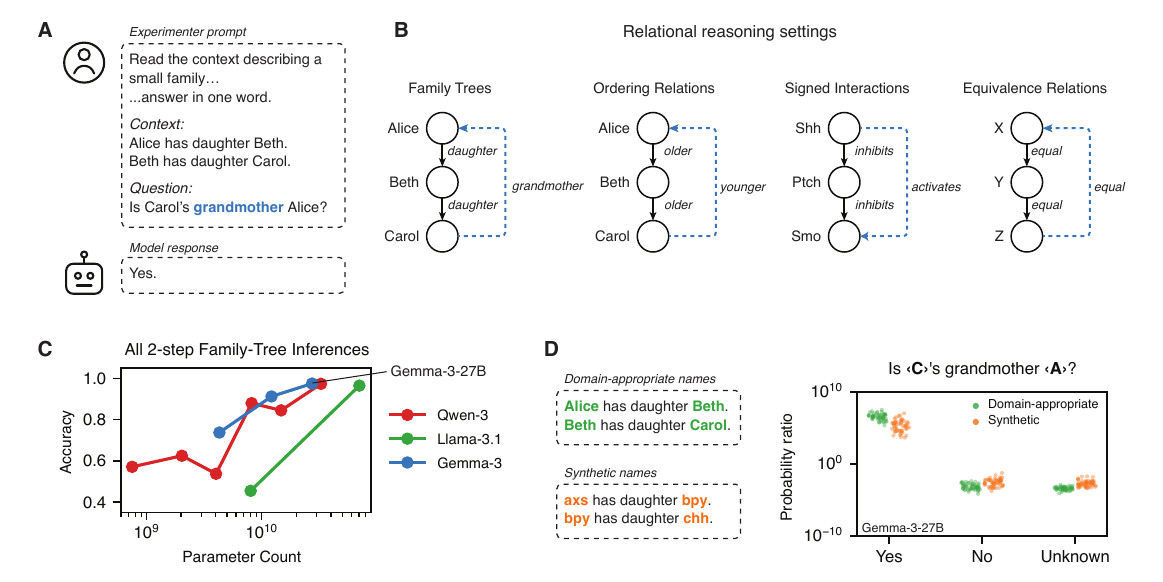}
    \caption{
\textbf{Behavioral task, model performance, and generalization.}
\textbf{A}, Task structure.
\textbf{B}, Relational reasoning settings. Solid black arrows denote examples of stated relationships; dashed blue arrows denote examples of implied relationships.
\textbf{C}, Performance in the family-tree setting. Colors denote model families; points denote individual models. Gemma-3-27B is highlighted here and is shown as an example in subsequent main-text figures; results across models and relational-reasoning settings appear in the SI.
\textbf{D}, Generalization to synthetic names. Green denotes domain-appropriate names (common English names); orange denotes synthetic names (random three-letter strings). Each condition shows 50 trials, with entity names resampled on each trial. ‹A› and ‹C› denote the first and third entity names on each trial. Each trial contributes one point per plotted answer token. Probability ratios are next-token probabilities divided by the geometric mean across the plotted answer tokens.
    }
    \label{fig:behavior}
\end{figure*}

%% file: manuscript/sections/results_01_behavior.tex
\subsection{Large language models infer unstated relationships and generalize to novel entity names}

To study how language models organize newly presented information for inference, we designed a set of relational-reasoning tasks that require models to answer questions about relationships implied by a short prompt. On each trial, models received a context stating two relationships among three entities, then were asked a question about a relationship determined by that context (Fig.~\figref{fig:behavior}A). We used no task-specific fine-tuning or in-context demonstrations. Models were evaluated across 14 relational-reasoning settings spanning four task families: family trees, ordering relations (e.g., age comparisons), signed interactions (e.g., biological interaction networks), and equivalence relations (Fig.~\figref{fig:behavior}B; SI Sec.~\ref{SI-sec:prompt_templates}).

We studied the behavior of a collection of 11 open-weight transformer models spanning multiple model families (Gemma-3, Llama-3.1, Qwen-3) and scales ($\sim$600M to $\sim$70B parameters) on these tasks. Across most relational reasoning settings, sufficiently large models reliably answered questions about various unstated relationships implied by the context (Figs.~\figref{fig:behavior}C;~\ref{SI-fig:behavior-family-tree-yes-no}--\ref{SI-fig:behavior-variable-equivalence}). Furthermore, performance was preserved when entity names were replaced with synthetic names that lacked domain-appropriate semantic associations (Figs.~\figref{fig:behavior}D;~\ref{SI-fig:behavior-family-tree-yes-no}--\ref{SI-fig:behavior-variable-equivalence}). Together, these observations demonstrate a capacity for inference that is both flexible (i.e., models could infer a range of unstated relationships as required by the query) and systematic \cite{fodor1988connectionism} (i.e., the capacity for inference extended immediately to corresponding inferences about novel entities arranged in the same relational structure).

%% file: manuscript/sections/results_02_transformers.tex
\subsection{Probing internal representations in transformer models}

We next asked about the structure of the internal representations that support this capacity. We begin by identifying the relevant internal states in a transformer-based language model \cite{vaswani_2017_transformer} that carry information about the stated relationships, then describe our approach to probing their organization.

To process text, transformer models first divide the text into tokens, associate an initial vector with each token, then update these vectors through a sequence of layers (Fig.~\figref{fig:separability}A). We collect the vectors associated with the relationship statements into an object we call the \textit{activation tensor}:
\begin{equation}
\A
=
\Big\{
\mathbf{h}_{tl}
\Big\}
_{\begin{subarray}{l}
t=t_{\mathrm{start}},\ldots,t_{\mathrm{end}} \\
l=1,\ldots,N_{\mathrm{layers}}
\end{subarray}},
\end{equation}
where $\mathbf{h}_{tl}$ denotes the vector generated at token position $t$ and layer $l$, and $t_{\mathrm{start}}$ and $t_{\mathrm{end}}$ delimit the relevant segment of the prompt. 

Within each layer, the self-attention mechanism permits the retrieval of information from earlier token positions. Retrieval is mediated by key--value pairs generated from each vector by learned linear projections. The keys determine which pairs are accessed, while the values supply the retrieved information. We collect the key--value pairs associated with the relationship statements into an object we call the \textit{key--value memory}:
\begin{equation}
\M
=
\Big\{
(\mathbf{k}_{tl}^{(i)},
 \mathbf{v}_{tl}^{(i)})
\Big\}_{\begin{subarray}{l}
t=t_{\mathrm{start}},\ldots,t_{\mathrm{end}} \\
l=1,\ldots,N_{\mathrm{layers}} \\
i=1,\ldots,N_{\mathrm{KV}}
\end{subarray}},
\end{equation}
where $(\mathbf{k}_{tl}^{(i)},\mathbf{v}_{tl}^{(i)})$ denotes the $i$th key--value pair at token position $t$ and layer $l$, and $N_{\mathrm{KV}}$ is the number of key--value pairs generated per vector. 

Crucially, the dependence of the models' responses on the stated relationships is mediated entirely through $\A$ and $\M$. Specifically, information about the relationships is encoded in $\A$, then made available for retrieval through the key--value memory $\M$:
\begin{equation}
\text{relationships}
\longrightarrow \A
\longrightarrow \M
\xrightarrow{\mathrm{query}}
\text{response}.
\end{equation}
Questions about the structure of the cognitive map can therefore be operationalized as questions about the organization of information in $\A$ and $\M$.

To probe how information is organized within these objects, we designed a family of in silico experiments---here termed \textit{mosaic experiments}---based on the approach described in \cite{feng_2024_entity_binding}. In each experiment, we constructed a \textit{mosaic activation tensor} by recombining parts of two or more \textit{source activation tensors} generated as the model processed different prompts. Mosaic key--value memories were constructed in the same way. For example, in Fig. \figref{fig:separability}B, we recorded activation tensors for two prompts describing the same relational structure instantiated with different entity names, then constructed $\Amosaic$ by replacing the row of $\Actrl$ corresponding to the first “Beth” token with the corresponding row for “Eve” in $\Adonor$. We then presented the model with a query while holding the internal states corresponding to $\A$ or $\M$ fixed at the values specified by the mosaic; all other internal states were computed normally. The model’s responses under such a manipulation provide a behavioral readout of the relationships supported by the mosaic. As we show next, different hypotheses about the organization of information in $\A$ or $\M$ make different predictions about the outcomes of such experiments.

%% file: manuscript/figures/fig_2_caption.tex
\begin{figure*}[p]
    \centering
    \includegraphics[width=\textwidth]{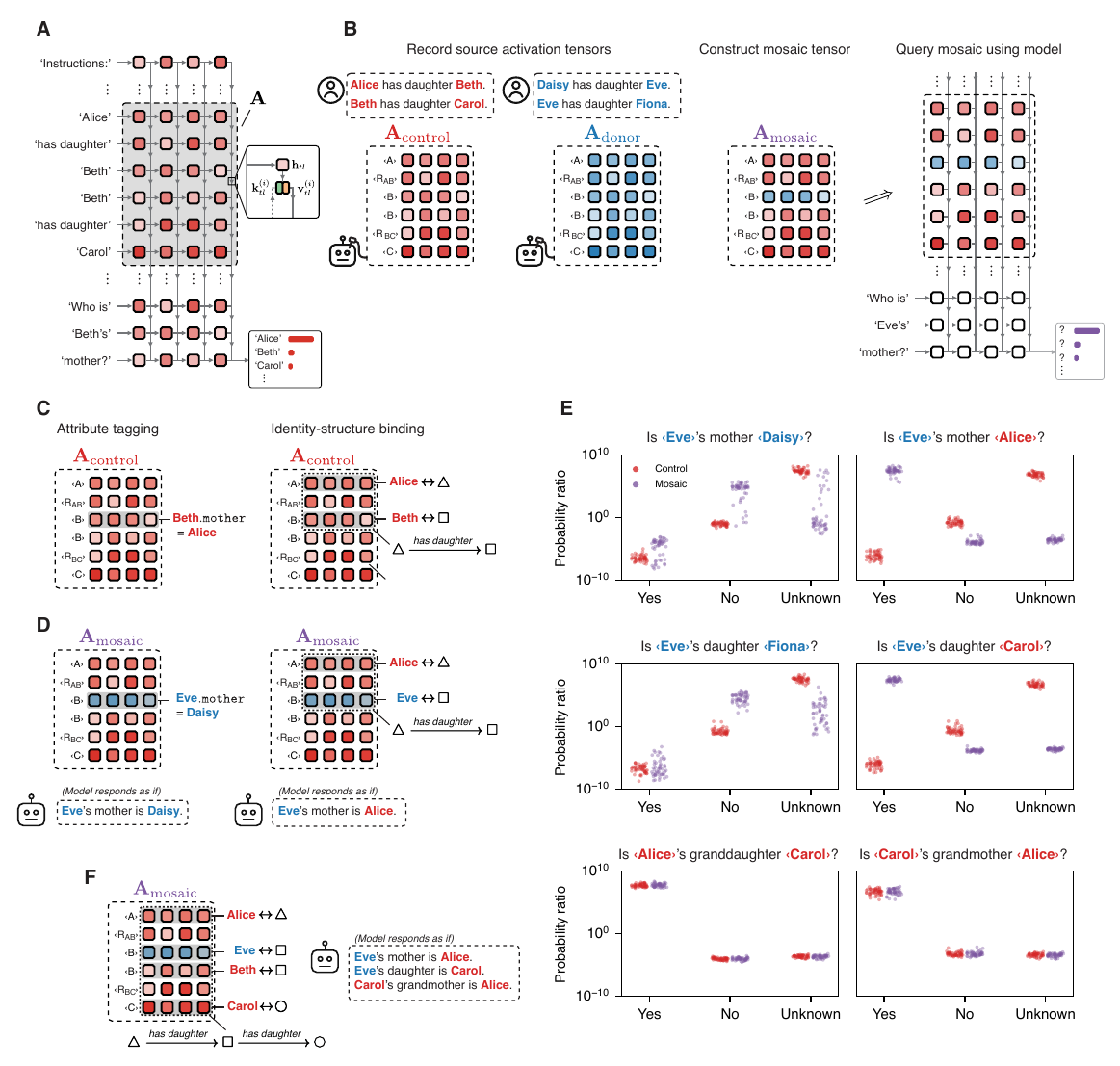}
    \caption{
\textbf{Mosaic experiments probe the organization of information within the activation tensor.}
\textbf{A}, Information flow through a transformer. Tokenization is schematic in this and subsequent diagrams. Squares represent vectors; the dashed box encloses the activation tensor. Arrows indicate the direction of causal dependence. Later vectors retrieve information through keys (green) and values (orange) generated from earlier vectors (inset).
\textbf{B}, Schematic of a mosaic experiment. Source tensors are recorded and recombined into a mosaic tensor. The model then processes a query with relevant internal states held fixed at their mosaic values; subsequent activations (unfilled squares) are computed normally. Next-token probabilities provide a readout of the relationships supported by the mosaic.
\textbf{C}, Competing hypotheses for the representation of a relationship in the activation tensor.
\textbf{D}, Predicted outcomes under each hypothesis.
\textbf{E}, Model responses with control (red) and mosaic (purple) tensors; 50 trials per condition. Probability ratios are defined in Fig. \figref{fig:behavior}D; names in query headings are illustrative. Results are shown for Gemma-3-27B in the family-tree setting; see SI for results across other models and relational-reasoning settings.
\textbf{F}, Inferred organization. Symbols (denoted by shapes) label roles in the full relational structure; entity names are bound to these roles at corresponding token rows.
    }
    \label{fig:separability}
\end{figure*}

%% file: manuscript/sections/results_03_separability.tex
\subsection{Relational structure is represented separately from entity identity}

We first asked how a single relationship (e.g., ``Alice has daughter Beth'') is encoded in the activation tensor $\A$. We considered two hypotheses:

Under the \emph{attribute tagging hypothesis}, models encode relationships as an attribute of one of the participating entities (Fig. \figref{fig:separability}C, left). This hypothesis is motivated by empirical evidence indicating that language models express facts learned during training by tagging representations at entity tokens with features that encode attributes of those entities \cite{meng_2022_factual_associations, hernandez_2023_relation_decoding, geva_2023_factual_recall}. Analogously, relationships introduced in the prompt could be encoded by constructing such features on the fly. For example, the models could encode the relationship ``Alice has daughter Beth'' by adding to the representations at ``Beth'' a feature corresponding to the attribute ``has mother Alice'' (e.g., using a conjunctive code \cite{barak_2013_mixed_selectivity, frankland_2026_curse_of_generalization} for ``has mother'' and ``Alice''). Because the complete relationship is not available until the latter entity name (e.g., ``Beth'') is processed, such a feature is constrained to appear in the row of $\A$ corresponding to the latter entity token.

An alternative hypothesis, the \emph{identity--structure binding hypothesis}, is that models carry an abstract, name-independent representation of the relationship, then bind the names to that representation (Fig.~\figref{fig:separability}C, right). This hypothesis is related to recent proposals that language models can learn to perform computations with properties traditionally associated with symbolic processing \cite{feng_2024_entity_binding, prakash_2024_finetuning_mechanisms, yang_2025_symbolic_mechanisms, assouel_2025_visual_symbols, cui_2026_spatial_binding, prakash_2025_belief_lookbacks}. Under this hypothesis, changing the entity names in the prompt should leave the abstract representation of the relationship intact while altering the name-specific information bound to that representation.

The mosaic experiment illustrated in Fig. \figref{fig:separability}B was designed to distinguish between the two hypotheses. In this experiment, $\Amosaic$ was constructed from two source tensors, $\Actrl$ and $\Adonor$, recorded as models processed two contexts describing the same relational structure instantiated with different sets of entity names. For example, the control and donor contexts could read ``Alice has daughter Beth; Beth has daughter Carol'' and ``Daisy has daughter Eve; Eve has daughter Fiona'', respectively. $\Amosaic$ would be constructed by replacing the row in $\Actrl$ corresponding to the first occurrence of ``Beth'' with the corresponding row for ``Eve'' in $\Adonor$.

The two hypotheses make different predictions for the outcome of the experiment. The attribute tagging hypothesis predicts that the substituted row in $\Amosaic$ carries the feature ``has mother Daisy''. Models should therefore behave as if Eve's mother were Daisy (Fig.~\figref{fig:separability}D, left). In contrast, the identity--structure binding hypothesis predicts that replacing Beth's row in $\Actrl$ with the corresponding row from $\Adonor$ leaves a name-independent representation of the abstract relationship intact while changing the name bound to that representation from ``Beth'' to ``Eve''. Models should therefore behave as if Eve's mother were Alice (Fig.~\figref{fig:separability}D, right). Note that we use the names ``Alice'', ``Beth'', \ldots, and the family tree setting for illustrative purposes; for each experiment we present, we resampled entity names on each trial and performed analogous experiments for all relational reasoning tasks. 

Consistent with the identity--structure binding hypothesis, we found that models could recover the relationship ``Eve has mother Alice'', but not ``Eve has mother Daisy'', from $\Amosaic$ (Fig.~\figref{fig:separability}E, top). Unexpectedly, we also found that models could recover the relationship ``Eve’s daughter is Carol'' (Fig.~\figref{fig:separability}E, middle), and the two-step relationship ``Alice’s granddaughter is Carol'' and its inverse (Fig. \figref{fig:separability}E, bottom). Thus, models recovered relationships from $\Amosaic$ as if Eve's name had become associated with Beth's role in the full control family tree. We observed similar results when models answered queries requiring an entity name, such as ``Who is Eve’s mother?'' (Fig.~\figref{SI-fig:query_format_who_is}). These patterns suggest a more general version of the identity--structure binding hypothesis, where the activation tensor binds a representation of each entity name to some component---common to $\Actrl$ and $\Adonor$---representing some feature of the prompt that is independent of entity identity.

What feature of the prompt does this component represent? We consider three possibilities. One possibility is that it represents the sequential structure of the prompt. Specifically, $\A$ could bind tokens to a sequence of position codes:
\begin{equation}
\begin{gathered}
\mathcal{P}_1, \ \mathcal{P}_2, \ \mathcal{P}_3, \ 
\mathcal{P}_4, \ \mathcal{P}_5, \ \mathcal{P}_6
\\
(\mathcal{P}_1 \leftrightarrow \mathrm{Alice}),
(\mathcal{P}_2 \leftrightarrow \text{has daughter}), \ldots .
\end{gathered}
\end{equation}
Under this account, $\Amosaic$ would encode the sequence ``Alice has daughter Eve; Beth has daughter Carol.'' This would explain why models treated Alice as Eve's mother (Fig. \figref{fig:separability}E, top); however, it is inconsistent with the models' belief that Carol is Eve’s daughter or Alice’s granddaughter (Fig. \figref{fig:separability}E, middle and bottom). As an additional test, we evaluated models on mosaic tensors designed to permute either the positional signals supplied by the transformer architecture or any internally constructed codes for global token position. Neither manipulation appreciably changed the models' behavior (Figs.~\figref{SI-fig:S1_rope_swap},~\figref{SI-fig:S2_indirect_swap}).

A second possibility is that the name-independent component of $\A$ represents two abstract mother--daughter relationships. Entity names could then be bound to the mother and daughter roles in each relationship:
\begin{equation}
\begin{gathered}
\mathcal{M}_1 \xrightarrow{\text{has daughter}} \mathcal{D}_1, \
\mathcal{M}_2 \xrightarrow{\text{has daughter}} \mathcal{D}_2
\\
(\mathcal{M}_1 \leftrightarrow \mathrm{Alice}), \
(\mathcal{D}_1 \leftrightarrow \mathrm{Beth}), \\
(\mathcal{M}_2 \leftrightarrow \mathrm{Beth}), \
(\mathcal{D}_2 \leftrightarrow \mathrm{Carol}).
\end{gathered}
\end{equation}
Because the two relationships are linked only through the shared name ``Beth'', the account likewise cannot explain the middle or bottom observations in Fig. \figref{fig:separability}E.

A third possibility is that the name-independent component of $\A$ represents the full relational structure of the prompt, including the fact that the daughter in the first clause and the mother in the second clause correspond to the same entity. The abstract structure represented in $\A$ can be depicted in terms of three \textit{symbols}, which stand for the three unique entities that participate in the relational structure:
\begin{equation}
\begin{gathered}
\triangle \xrightarrow{\text{has daughter}} \square, \ 
\square \xrightarrow{\text{has daughter}} \scaledcircle \\ 
(\triangle \leftrightarrow \mathrm{Alice}), \
(\square \leftrightarrow \mathrm{Beth}), \\
(\square \leftrightarrow \mathrm{Beth}), \
(\scaledcircle \leftrightarrow \mathrm{Carol}).
\end{gathered}
\end{equation}
This account explains all three of our observations and is diagrammed in Fig. \figref{fig:separability}F.

To test the generality of this representational organization, we performed analogous mosaic experiments across relational reasoning settings and models. Similar patterns held across every setting (Figs.~\figref{SI-fig:separability-gemma27b-family-tree-yes-no}--\figref{SI-fig:separability-gemma27b-variable-equivalence}) and model (Figs.~\figref{SI-fig:model-generalization-separability-gemma-4b}--\figref{SI-fig:model-generalization-separability-qwen-32b}) we tested, provided the model could perform the underlying relational reasoning task.

%% file: manuscript/figures/fig_3_caption.tex
\begin{figure*}[t]
    \centering
    \includegraphics[width=\textwidth]{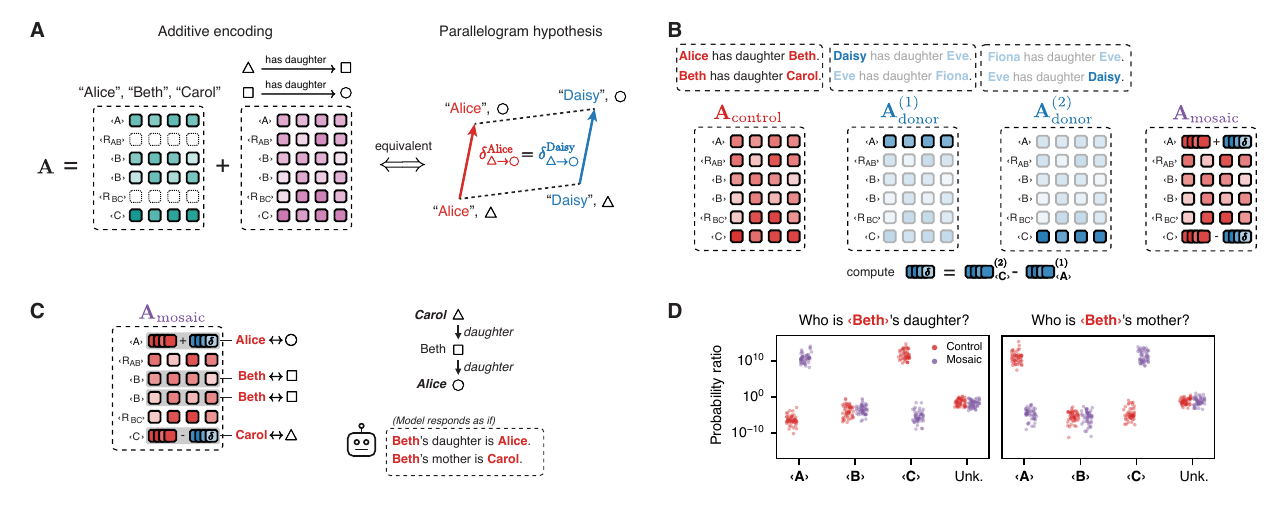}
    \caption{
\textbf{Representations of entity identity and role combine additively.}
\textbf{A}, Parallelogram hypothesis. The activation tensor is hypothesized to decompose into a name-independent component (pink) carrying relational structure and a name-dependent component (teal) carrying entity identity (left); geometrically, the four representations of any two names in any two roles form a parallelogram (right).
\textbf{B}, Mosaic experiment. Two donor prompts place the same entity name in different roles; subtracting the corresponding rows provides an estimate of the role-swap vector $\boldsymbol{\delta}$. Adding and subtracting this vector to and from selected rows in a control tensor produces the mosaic.
\textbf{C}, Predicted outcome under additive encoding. The model should behave as if two control entities had exchanged roles.
\textbf{D}, Model responses with control (red) and mosaic (purple) tensors; 50 trials per condition. Probability ratios are defined in Fig. \figref{fig:behavior}D; names in query headings are illustrative. ‹A›, ‹B›, and ‹C› denote the three entity names in the control context in order of first appearance. We show results for Gemma-3-27B in the family-tree setting; see SI for results across other models and relational-reasoning settings.
}
    \label{fig:additivity}
\end{figure*}

%% file: manuscript/sections/results_04_linearity.tex
\subsection{Representations of entity identity and relational structure combine additively}

We next asked how entity names are bound to their roles in the relational structure. One natural hypothesis is that the binding is mediated by the co-occurrence, within the same row of $\A$, of a pattern of activity representing an entity's name and a pattern of activity determining the role to which it is bound. The latter pattern of activity could then be regarded as the model's realization of the symbol standing for that entity. 

Under the simplest version of this hypothesis, which we call the \textit{parallelogram hypothesis}, $\A$ can be written as the sum of a name-dependent component carrying a representation of entity identity and a name-independent component carrying a representation of relational structure (Fig. \figref{fig:additivity}A, left). A row of $\A$ corresponding to the name $n$ bound to the symbol $s$, denoted $\mathbf{a}(n,s)$, would then decompose as:
\begin{equation}
\mathbf{a}(n, s) \approx \boldsymbol{\phi}_{\text{name}}(n) + \boldsymbol{\phi}_{\text{sym}} (s),
\end{equation}
where $\boldsymbol{\phi}_{\mathrm{name}}$ and $\boldsymbol{\phi}_{\mathrm{sym}}$ map entity names and symbols, respectively, to their corresponding patterns of activity. This hypothesis is equivalent to the geometric statement that the representations of all four pairings of any two entity names with any two roles form a parallelogram (Fig. \figref{fig:additivity}A, right). Analogous representational geometries underlie classical models of analogical reasoning \cite{rumelhart_1973_analogical_reasoning} and have been observed in learned word embeddings \cite{mikolov_2013_linguistic_regularities}.

The parallelogram hypothesis predicts that the change in activity associated with changing an entity's role should not depend on the name of that entity:
\begin{equation}
\mathbf{a}(n, s_2) - \mathbf{a}(n, s_1) \approx \boldsymbol{\phi}_{\text{sym}}(s_2) - \boldsymbol{\phi}_{\text{sym}}(s_1) \equiv \boldsymbol{\delta}_{s_1 \to s_2}.
\end{equation}
We refer to $\boldsymbol{\delta}_{s_1 \to s_2}$ as a \textit{role-swap vector}, because adding and subtracting $\boldsymbol{\delta}_{s_1 \to s_2}$ to and from the representations of any two entities bound to symbols $s_1$ and $s_2$ is predicted to make models behave as if the two entities had exchanged roles.

Fig. \figref{fig:additivity}B illustrates the design of a mosaic experiment that tests this prediction. We first recorded two donor tensors, $\Adonor^{(1)}$ and $\Adonor^{(2)}$, using two prompts that placed the same entity in two different roles. If the representations of name and role combine additively, subtracting the corresponding entity rows in $\Adonor^{(1)}$ and $\Adonor^{(2)}$ should cancel their shared name component, yielding an estimate of the role-swap vector. To build the mosaic tensor $\Amosaic$, we added and subtracted the estimated role-swap vector to and from the relevant rows of a control tensor, $\Actrl$, recorded from a prompt containing a different set of entity names. The parallelogram hypothesis predicts that this operation should exchange the role-dependent components of the two rows while leaving their name-dependent components unchanged, causing models to recover relationships from $\Amosaic$ as if the two control entities had exchanged roles (Fig. \figref{fig:additivity}C).

Models' responses were consistent with this prediction (Fig. \figref{fig:additivity}D). Similar patterns held across most relational-reasoning settings (Figs.~\figref{SI-fig:additivity-gemma27b-family-tree-yes-no}--\figref{SI-fig:additivity-gemma27b-variable-equivalence}) and models (Figs.~\figref{SI-fig:model-generalization-additivity-gemma-4b}--\figref{SI-fig:model-generalization-additivity-qwen-32b}) that we tested. These observations demonstrate that the activation tensor decomposes linearly into a name-independent component representing the relational structure and a name-dependent component representing the identities of the entities that instantiate it.

%% file: manuscript/figures/fig_4_caption.tex
\begin{figure*}[t]
    \centering
    \includegraphics[width=0.999\textwidth]{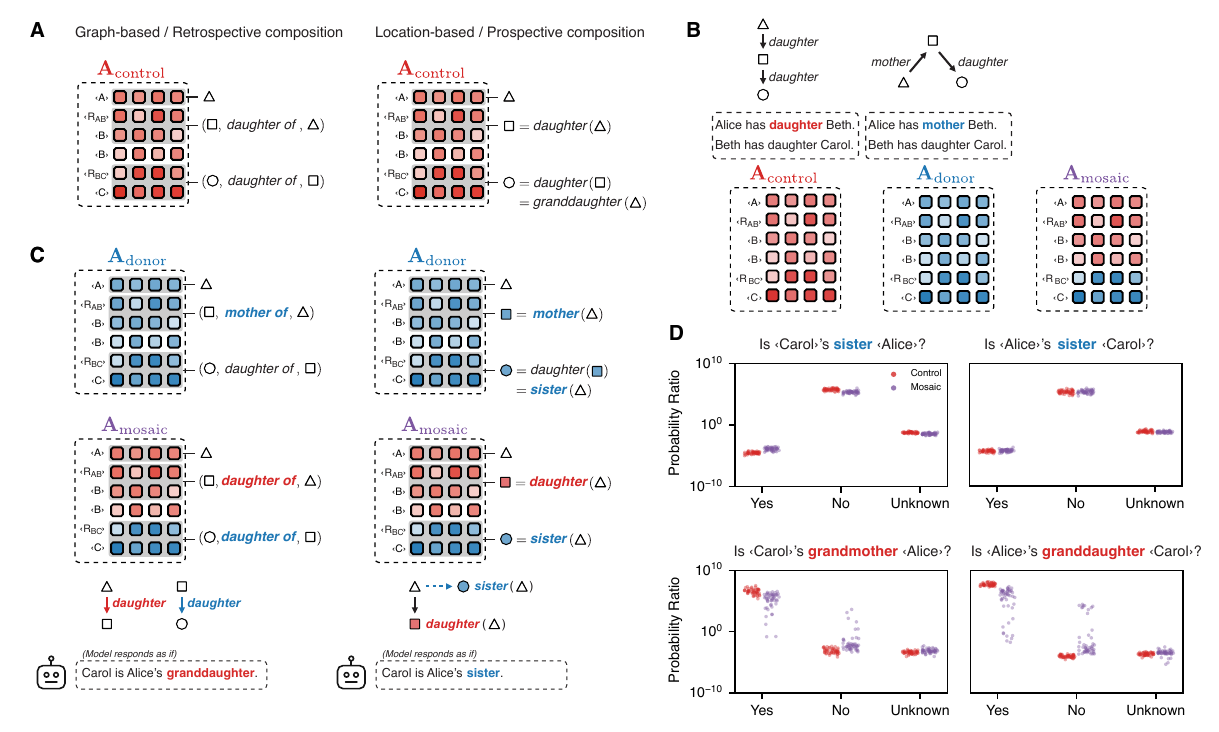}
    \caption{
\textbf{A mosaic experiment distinguishes retrospective and prospective composition of relationships.}
\textbf{A}, Competing hypotheses for the representation of relational structure in the activation tensor. Relationships could be stored as graph edges and composed retrospectively at query time (left), or composed prospectively as they are encountered to assign entities locations in an internal coordinate system (right).
\textbf{B}, Mosaic experiment. The control and donor prompts differ only in the identity of the first relationship. Rows from the start of the second relationship phrase through the final entity in the control tensor are replaced with the corresponding donor rows.
\textbf{C}, Predicted outcomes under each hypothesis.
\textbf{D}, Model responses with control (red) and mosaic (purple) tensors; 50 trials per condition. Probability ratios are defined in Fig. \figref{fig:behavior}D; names in query headings are illustrative. We show results for Gemma-3-27B in the family-tree setting; see SI for results across other models and relational-reasoning settings.
    }
    \label{fig:composition}
\end{figure*}

%% file: manuscript/sections/results_05_prospective_retrospective.tex
\subsection{Multi-step inference relies primarily on a graph-like representation of observed relationships}

We next asked how relational information is organized within the name-independent component of $\A$. We considered two hypotheses:

Under the \textit{retrospective composition hypothesis}, the name-independent component of $\A$ represents the relational structure as a labeled, directed graph (Fig. \figref{fig:composition}A, left). Symbols correspond to nodes in this graph, and each observed relationship corresponds to an edge between two nodes. Because multi-step relationships are represented implicitly as paths through the graph, this representational strategy requires models to retrospectively compose the relevant relationships once a query is asked. This computation is related to classical accounts of model-based planning over an internal world model \cite{shannon_1950_computer_chess, degroot_1965_thought_choice_chess, mattar_2026_planning}.

An alternative hypothesis, the \textit{prospective composition hypothesis}, is that models prospectively compose relationships as they are encountered so that the name-independent activity in $\A$ associated with each entity reflects the full sequence of relationships leading up to it (Fig.~\figref{fig:composition}A, right). Under this hypothesis, symbols would correspond to particular locations in an internal coordinate system, and relationships would correspond to transformations (e.g., translations or rotations) between those locations. Successively applying these transformations as relationships are encountered would determine the location of each new entity. If the transformations compose in the same way as the corresponding relationships, then multi-step relationships would be automatically reflected in the locations and would be recoverable at query time without retracing the intermediate relationships. This computation is directly analogous to path integration in models of animal spatial navigation \cite{burak_2009_path_integration, whittington_2020_tolman_eichenbaum_machine}; related schemes have also been proposed for representing small family trees \cite{paccanaro_2001_relational_embedding}.

To distinguish between these possibilities, we designed a mosaic experiment to test whether or not the task-relevant component of the name-independent activity associated with the final entity is dependent on the full history of preceding relationships (Fig.~\figref{fig:composition}B). In this experiment, $\Amosaic$ was constructed from two source tensors, $\Actrl$ and $\Adonor$, recorded as models processed two prompts that differed only in the identity of the first relationship. For example, the control and donor contexts could read ``Alice has daughter Beth; Beth has daughter Carol'' and ``Alice has \textit{mother} Beth; Beth has daughter Carol'', respectively. We then constructed $\Amosaic$ by replacing the rows associated with the final entity in $\Actrl$ with the corresponding rows from $\Adonor$ (Fig.~\figref{fig:composition}B).

Under the retrospective composition hypothesis, the substituted rows encode only the relationship that directly introduces the final entity. Thus, at query time, models should compose the first relationship from the control prompt with the second relationship from the donor prompt; the multi-step relationships supported by $\Amosaic$ should therefore match those implied by the \textit{control} prompt (Fig.~\figref{fig:composition}C, left). In contrast, under the prospective composition hypothesis, the substituted rows reflect the full history of preceding relationships and encode the final entity’s location as specified by the donor prompt. The multi-step relationships supported by $\Amosaic$ should therefore match those implied by the \textit{donor} prompt (Fig.~\figref{fig:composition}C, right).

Consistent with the retrospective composition hypothesis, we found that the multi-step relationships that models recovered from $\Amosaic$ generally matched those implied by the control prompt (Fig. \figref{fig:composition}D). Similar patterns held across relational reasoning tasks (Figs.~\figref{SI-fig:retrospective-prospective-family-tree-yes-no}--\figref{SI-fig:retrospective-prospective-variable-equivalence}) and models (Figs.~\figref{SI-fig:model-generalization-retrospective-prospective-gemma-4b}--\figref{SI-fig:model-generalization-retrospective-prospective-qwen-32b}). However, in a subset of conditions, the models' responses deviated systematically from this pattern, shifting toward---and in some cases closely matching---the responses implied by the donor prompt (Figs.~\figref{SI-fig:S3_prospective_retrospective_exceptions}, \figref{SI-fig:model-generalization-retrospective-prospective-llama-70b}D). These deviations were most apparent in the family-tree setting.

Overall, our observations suggest that models rely primarily on a graph-like representation of the observed relationships in $\A$ and compose the relevant relationships retrospectively when a query is asked. However, in certain settings, models may also rely on other sources of information in $\A$ that reflect the prospective composition of observed relationships. Consistent with this possibility, directly visualizing representations at different token positions revealed a history-dependent component at relation tokens (Fig.~\figref{SI-fig:family_tree_pca}). Why models rely on such information for certain queries but not others remains unclear. One possibility is that the variation reflects some statistical feature of the training data, as suggested for other behavioral idiosyncrasies in language models \cite{mccoy_2024_embers_autoregression}.

%% file: manuscript/figures/fig_5_caption.tex
\begin{figure*}[t]
    \centering
    \includegraphics[width=\textwidth]{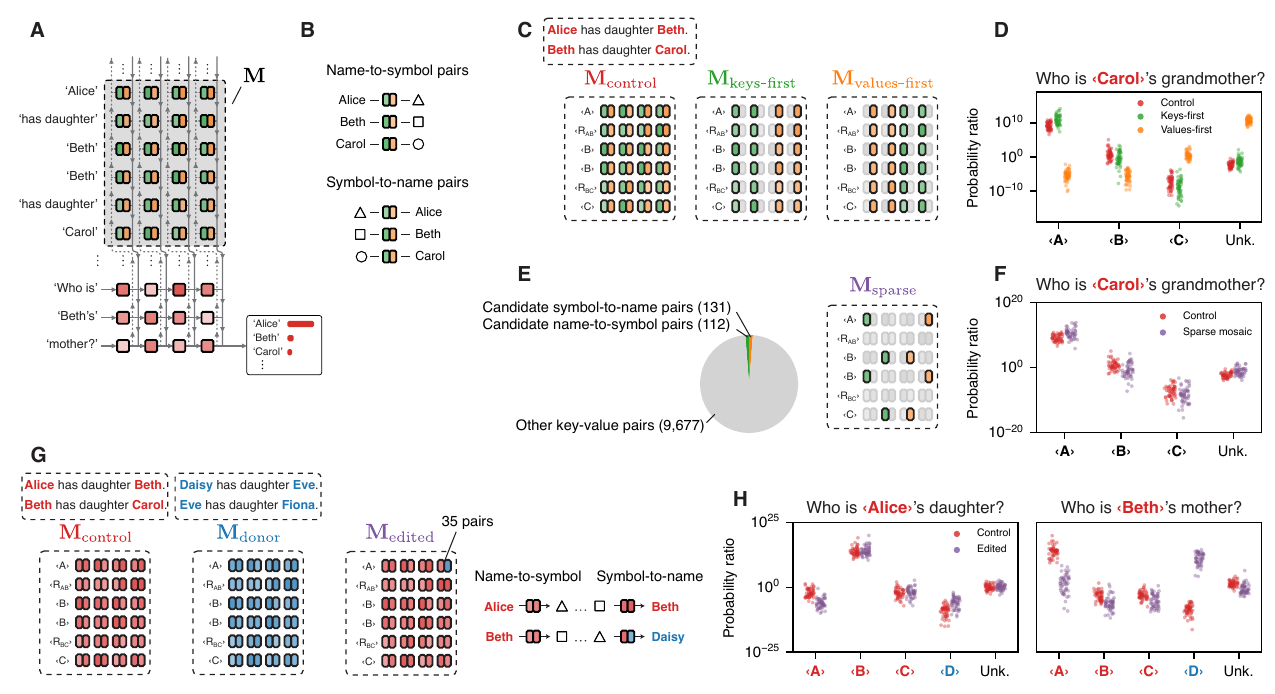}
    \caption{
\textbf{Identifying and editing key--value pairs linking entity names and roles.}
\textbf{A}, Key--value memory and information flow through a transformer. Linear projections generate keys (green) and values (orange) from the activation tensor; one pair per layer and position is shown. Upward dotted arrows indicate queries; downward arrows indicate retrieved values.
\textbf{B}, Hypothesized name-to-symbol and symbol-to-name pairs.
\textbf{C}, Keys-first and values-first mosaic memories. Grey shading indicates replacement by averages across prompts with resampled names.
\textbf{D}, Model responses with the control memory (red), keys-first (green), and values-first (orange) mosaics.
\textbf{E}, Pruned memory with name dependence confined to a sparse subset of key–value pairs.
\textbf{F}, Model responses with the control memory (red) and sparse mosaic memory (purple).
\textbf{G}, Editing a symbol-to-name association. Values of candidate symbol-to-name pairs for one control entity are replaced with the corresponding donor values (left). This should alter the model's answer to one query but not its inverse (right).
\textbf{H}, Model responses with the control memory (red) and edited memory (purple). ‹A›–‹C›: control names; ‹D›: first donor name. D, F, H use the same plotting conventions as elsewhere. Results are shown for Gemma-3-27B in the family-tree setting; see SI for results across other models and relational-reasoning settings.
    }
    \label{fig:binding}
\end{figure*}

%% file: manuscript/sections/results_06_key_value_binding.tex
\subsection{Separate sets of key--value pairs support name-to-symbol and symbol-to-name retrieval}

Our preceding findings suggested that $\A$ decomposes into two components, one representing relational structure and the other entity identity. We next asked how models use each kind of information to access the other. To address this question, we studied the organization of information in the key--value memory $\M$, from which the models directly retrieve information (Fig. \figref{fig:binding}A).

Recall that $\M$ consists of a large set of key--value pairs constructed from $\A$ via linear projection. Together with our finding that representations of names and symbols combine linearly in $\A$, this suggests a natural hypothesis for how information is organized in $\M$ to support retrieval of an entity’s symbol from its name and vice versa. Specifically, certain linear projections could produce key--value pairs in $\M$ for which the key carries information about an entity's name and the value carries information about its symbol, supporting retrieval of an entity's symbol from its name. We refer to these hypothesized key--value pairs as \textit{name-to-symbol pairs}. A separate set of \textit{symbol-to-name pairs} with symbol information in the keys and name information in the values could support retrieval in the reverse direction (Fig. \figref{fig:binding}B).

The hypothesis that the model relies on these two kinds of key--value pairs to answer queries makes two key predictions. First, information dependent on entity identity should be required only in a subset of entries in $\M$. Specifically, name-dependent information is required in the keys but not in the values of name-to-symbol pairs; the reverse is true of symbol-to-name pairs. Second, name-to-symbol pairs should be accessed before symbol-to-name pairs. For example, to answer ``Who is Beth's mother?'', models must first retrieve Beth's symbol from her name, then identify her mother's symbol and retrieve the name bound to it.

To test these predictions, we constructed mosaic key--value memories designed to restrict where name-dependent information was available to the model. Specifically, we replaced selected keys or values with their averages across prompts where the names had been resampled. In $\M_{\mathrm{keys-first}}$, name-dependent information was available only in the keys from the first half of $\M$ and in the values from the second half of $\M$; $\M_{\mathrm{values-first}}$ used the reverse arrangement (Fig.~\figref{fig:binding}C). The two predictions above imply that models should be able to answer queries using $\M_{\mathrm{keys-first}}$, but not $\M_{\mathrm{values-first}}$.

Consistent with this prediction, substituting $\M_{\mathrm{keys-first}}$ for the models' key--value memory preserved performance, whereas substituting $\M_{\mathrm{values-first}}$ did not (Fig. \figref{fig:binding}D). Similar patterns held across relational reasoning tasks (Figs.~\figref{SI-fig:erasure-family-tree-yes-no}A--\figref{SI-fig:erasure-variable-equivalence}A) and models (Figs.~\figref{SI-fig:model-generalization-erasure-gemma-4b}A--\figref{SI-fig:model-generalization-erasure-qwen-32b}A). These observations establish the existence of a set of name-to-symbol and symbol-to-name pairs sufficient to support the models' behavior.

We next sought to localize these pairs within $\M$. Starting from $\M_{\mathrm{keys-first}}$, we greedily averaged out name-dependent information from the remaining keys and values, selecting at each step the replacement estimated to have the smallest effect on behavior (Fig.~\figref{SI-fig:pruning_trajectories}A,B). The resulting mosaic preserved the models' behavior while confining name dependence to a very sparse subset of key--value pairs (Fig.~\figref{fig:binding}E,F), distributed broadly across layers (Fig.~\figref{SI-fig:pruning_trajectories}C). By construction, the surviving pairs retained name-dependent information only in their keys or only in their values. We therefore interpreted them as candidate name-to-symbol and symbol-to-name pairs, respectively. 

Finally we asked if, by modifying these candidate pairs, we could plant specific ``false memories'' in $\M$. If the candidate pairs mediate name--symbol binding as we have proposed, then answering a particular query (e.g., ``Who is Alice's daughter?'') and its inverse (e.g., ``Who is Beth's mother?'') should rely on disjoint subsets of the candidate pairs. It should therefore be possible to plant a ``unidirectional false memory'' in $\M$ that is expressed only when a relationship is queried in a particular direction.

We demonstrated this phenomenon in the control--donor paradigm from Fig.~\figref{fig:separability}. We began by recording two source memories, $\Mctrl$ and $\Mdonor$, as models processed two prompts with the same relational structure but different entity names. We then edited the candidate symbol-to-name pairs associated with ``Alice'' by replacing their values in $\Mctrl$ with the corresponding values from $\Mdonor$, leaving all other entries unchanged (Fig. \figref{fig:binding}G). We found that under this manipulation, models continued to identify Beth as Alice's daughter, but identified Daisy, rather than Alice, as Beth's mother (Fig. \figref{fig:binding}H). Thus, we planted false memories in $\M$ that exhibited an unusual property: the relationships recovered by the models could not be reconciled with a single, logically consistent family tree. Analogous manipulations altered models' responses in predictable ways across relational reasoning tasks (Figs.~\figref{SI-fig:erasure-family-tree-yes-no}--\figref{SI-fig:erasure-variable-equivalence}, panels D,E) and models (Figs.~\figref{SI-fig:model-generalization-erasure-gemma-4b}--\figref{SI-fig:model-generalization-erasure-qwen-32b}, panels D,E).

%% file: manuscript/sections/discussion.tex
In Fig. \figref{fig:cognitive-map}A, we present a schematic of our findings concerning the general organization of the cognitive map instantiated by $\A$ and $\M$. The name-independent components of $\A$ and $\M$ carry an abstract representation of the relational structure expressed by the prompt. This representation can be described in terms of a graph whose nodes are labeled by a set of abstract symbols and whose edges represent observed relationships between entities. Concrete entity names are bound to the corresponding symbols by vector addition in $\A$ and by a sparse but distributed set of key--value pairs in $\M$.

A procedure for inference over such a map is shown in Fig. \figref{fig:cognitive-map}B. This procedure has three stages: first, the symbol bound to the queried entity’s name is retrieved; second, the graph is traversed to identify the symbol corresponding to the answer entity; and third, the name bound to that symbol is retrieved. Because corresponding inferences about different sets of entities depend on the same intermediate computation, this procedure naturally generalizes to novel sets of entities (cf. Fig. \figref{fig:behavior}D).

We note that the system diagrammed in Fig. \figref{fig:cognitive-map} exhibits the defining features of a classical symbol-processing machine \cite{fodor1988connectionism}, namely, a reliance on structured, symbolic internal representations, and the use of operations over those representations that respect their combinatorial structure. Therefore, viewed in terms of the longstanding gap between connectionist and symbolic accounts of cognition \cite{smolensky_1988_proper_treatment,fodor_1990_systematicity,chalmers_1993_compositionality}, our results provide a striking example of a connectionist system---a modern large language model---that learns to approximate the operation of a classical symbol-processing machine (for related examples, see \cite{touretzky_1988_connectionist_production_system,smolensky_1990_tensor_product_binding,shastri_1993_systematic_reasoning, feng_2024_entity_binding, prakash_2024_finetuning_mechanisms, yang_2025_symbolic_mechanisms, assouel_2025_visual_symbols, cui_2026_spatial_binding, prakash_2025_belief_lookbacks, campbell2024humanlikegeometricabstractionlarge}). On this view, our findings can be stated precisely as follows: the models' computations in our setting admit an \textit{approximate, high-level} description in terms of the symbol-processing machine diagrammed in Fig. \figref{fig:cognitive-map}. Although we have not explored the implications here, we emphasize that this description should be understood as approximate: indeed, we suspect that much of the flexibility and fluidity of cognitive systems can be understood in terms of lower-level departures from such symbolic descriptions.

In contrast to certain neuro-symbolic approaches that achieve abstract reasoning by building in symbolic structure by design \cite{mao2019nscl, ellis2021dreamcoder, manhaeve2018deepproblog}, the organization we describe here arises through learning without the need to prespecify any primitive operations or representations that are recognizably symbolic. What conditions, then, favor its emergence? Given the minimal set of operations built into the transformer architecture, our findings suggest that the following ingredients may be sufficient: first, a data distribution exhibiting certain symmetries (e.g., regularities in language that hold across changes in entity identity); second, a mechanism for flexibly binding together otherwise distinct representations to form associations relevant to the current context (e.g., the key--value structure of self-attention); and third, a learning mechanism driven by prediction error on these data (e.g., gradient descent on a next-token prediction objective). 

\input{manuscript/figures/fig_6_caption}

In the mammalian hippocampus and nearby structures, striking patterns of neural activity observed during spatial navigation \cite{okeefe_1971_place_cells, okeefe_1978_hippocampal_map, hafting_2005_grid_cells} and related tasks \cite{neupane_2024_mental_navigation, constantinescu_2016_conceptual_grid} have long been interpreted as signatures of a cognitive map. It has been proposed that these signatures reflect a map that separates the representation of structural and sensory information \cite{whittington_2020_tolman_eichenbaum_machine, chandra_2025_vectorhash, kymn2024binding}. Similar ingredients may be at work here as in language models: spatial navigation involves exploiting geometric regularities that recur across environments, and mammalian hippocampal circuitry has been proposed to supply a mechanism for rapidly and flexibly binding information from distinct processing systems \cite{cohen_1993_hippocampal_memory, teyler_1986_hippocampal_indexing, gershman_2025_key_value_memory}. These parallels may provide an example of how artificial and biological systems, despite profound differences in their physical substrates, nonetheless converge on similar computational principles.

Like many other capabilities of language models \cite{srivastava_2023_bigbench, wei_2022_emergent}, performance on our tasks improved markedly with model size. Beyond increasing model size, reasoning performance can also be improved by allowing models to generate intermediate reasoning tokens \cite{nye_2022_scratchpads, wei_2022_chain_of_thought}. What accounts for these improvements? If inference resembles retrospective search over a graph-like representation, both could provide greater capacity for the intermediate computations required for such a search. Beyond this, reasoning may also change the representation over which inference is performed. Here, we have treated the intermediate states produced as the model processes the prompt as the substrate of the cognitive map, because these states make information in the context available to influence later processing. Crucially, the same is true of the intermediate states produced as the model generates reasoning tokens. Therefore, once models are allowed to reason, the distinction between constructing a cognitive map and performing inference over it becomes blurred: models may dynamically refine and restructure the cognitive map as inference unfolds.

%% file: manuscript/figures/fig_6_caption.tex
\begin{figure}[t]
    \centering
    \includegraphics[width=0.4\textwidth]{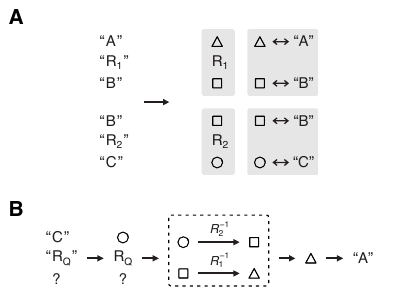}
    \caption{
\textbf{Structure and function of the cognitive map.}
\textbf{A}, General organization of the map.
Shapes denote abstract symbols for entities.
Vertical symbol--relation--symbol triples denote graph edges (left);
$\leftrightarrow$ denotes a name--symbol binding (right).
\textbf{B}, Inference over the map.
$\mathrm{R_Q}$ denotes the queried relation, here a two-hop relationship
(e.g., ``Who is Carol's grandmother?'').
The dotted rectangle encloses the portion of the computation
that is independent of entity identity.
    }
    \label{fig:cognitive-map}
\end{figure}